\pdfoutput=1

\documentclass[11pt]{article}

\usepackage[final]{coling}

\usepackage{times}
\usepackage{latexsym}

\usepackage[T1]{fontenc}

\usepackage[utf8]{inputenc}

\usepackage{microtype}

\usepackage{inconsolata}

\usepackage{graphicx}
\usepackage{times}
\usepackage{latexsym}
\usepackage{stmaryrd}
\usepackage{enumitem}
\usepackage{booktabs}
\usepackage{array} 
\usepackage[most]{tcolorbox}
\usepackage{bm}
\usepackage{subfigure}
\usepackage{subcaption}
\usepackage[utf8]{inputenc}
\usepackage{lipsum}
\usepackage{microtype}
\usepackage{bbm}
\usepackage{graphicx}
\usepackage{hyperref}
\usepackage{csquotes}
\usepackage{xcolor}
\usepackage{graphicx}
\usepackage{booktabs}
\usepackage{caption}
\usepackage{colortbl}
\usepackage{xcolor}
\usepackage{adjustbox}
\usepackage{makecell}
\usepackage{multirow}
\usepackage{amsmath}
\usepackage{amssymb}
\usepackage{inconsolata}
\usepackage{pifont} 

\usepackage{algorithm}
\usepackage[noend]{algorithmic}
\usepackage{amsmath}
\usepackage{fourier}

\usepackage{booktabs}
\usepackage{array}

\newcommand{\cmark}{\text{\ding{51}}}%
\newcommand{\xmark}{\text{\ding{55}}}%

\title{Towards Data Contamination Detection for Modern Large Language Models: Limitations, Inconsistencies, and Oracle Challenges}

\author{
 \textbf{Vinay Samuel\textsuperscript{1}\thanks{Authors contributed equally to this work.}\thanks{Correspondence: \href{mailto:vsamuel@andrew.cmu.edu}{vsamuel@andrew.cmu.edu}}},
 \textbf{Yue Zhou\textsuperscript{2}\footnotemark[1]},
 \textbf{Henry Peng Zou\textsuperscript{2}}
\\
 \textsuperscript{1}Carnegie Mellon University,
 \textsuperscript{2}University of Illinois Chicago
}

\begin{document}
\maketitle
\begin{abstract}
As large language models achieve increasingly impressive results, questions arise about whether such performance is from generalizability or mere data memorization. Thus, numerous data contamination detection methods have been proposed. However, these approaches are often validated with traditional benchmarks and early-stage language models, leaving uncertainty about their effectiveness when evaluating state-of-the-art ones with more challenging benchmarks. To address this gap and provide a dual investigation of the contamination status of state-of-the-art large language models and detection method robustness, we evaluate five contamination detection approaches with four state-of-the-art large language models across eight challenging and prevailing datasets. Our analysis reveals that (1) Current methods have non-trivial limitations in their assumptions and practical applications; (2) Notable difficulties exist in detecting contamination introduced during instruction fine-tuning with answer augmentation; and (3) Limited consistencies between state-of-the-art contamination detection techniques. These findings highlight the complexity of contamination detection in advanced language models and the urgent need for further research on robust and generalizable contamination evaluation. Our code is available at \hyperlink{https://github.com/vsamuel2003/data-contamination}{https://github.com/vsamuel2003/data-contamination}

\end{abstract}

\section{Introduction}

While large language models (LLMs) consistently achieve higher state-of-the-art results across various benchmarks~\citep{LLM-eval2, LLM-eval1, LLM-eval3, LLM-eval4}, the lack of curation in and the limited disclosure of massive training datasets raise a critical question: Does the model performance arise from model generalizability or mere memorization? Furthermore, were the test sets possibly contaminated without notice? These questions have become crucial in accurately gauging LLMs' performance and has led to a critical area of research: detecting data contamination in LLMs. 

Data contamination occurs when test or evaluation data is exposed to the model during its training phases (either pre-training or fine-tuning). This can lead to artificially inflated performance metrics through memorization rather than true generalization. 
Recent work in detecting data contamination in LLMs has primarily focused on detecting contamination through validating the log probability of the data in the datasets \citep{contaminat-baseline-canonical-p-value-ICLR24-oral, contaminat-baseline-min-k-prob-AUC-ICLR24} or determining contamination through prompting-based approaches \citep{contaminat-baseline-completion-overlap-ICLR24-spotlight, contaminat-baseline-perturb-quiz}. Additional studies have been aimed at understanding the different types of contamination, such as differentiating between training or development split contamination and testing set contamination \citep{sainz2023nlp}  and contamination occurring in the pertaining phase versus the supervised fine-tuning stage \citep{jacovi2023stop}.  

However, existing research in this area has several limitations. \textit{Firstly}, most methods are shown to be effective on traditional benchmarks, which are likely overexposed online and to LLMs. In contrast, challenging datasets that test the limits of LLM capabilities are neglected. These newer benchmarks are often more complex with novel formats. \textit{Secondly}, the tested LLMs are often early-staged ones, such as GPT-J, while the rapid pace of LLM development has left a gap in understanding contamination in the latest models. \textit{Thirdly}, previous research predominantly focuses on (possibly unintentional) contamination occurring during pretraining, where models are exposed to data in its original form. However, it overlooks contamination during instruction fine-tuning, where original data is subtly modified by, for instance, answer augmentation with chain-of-thought reasoning steps. Such variations may cause difficulties in contamination detection yet are rarely considered when evaluating the effectiveness of contamination detection methods. \textit{Lastly}, while various detection methods have demonstrated effectiveness (with a particular set of datasets and LLMs of their choice), there is a lack of a comprehensive cross-comparison to assess the consistency and reliability of these techniques, particularly for more recent LLMs, benchmarks, and training paradigms.

To bridge these gaps, we evaluate five distinct data contamination detection approaches, including three state-of-the-art methods recently published at ICLR, a simple prompting method based on token perturbation, and our proposed pilot prompt-based method, which queries the LLM about its knowledge of the original order of data points. Our study covers eight benchmarks, including six challenging datasets frequently used to evaluate modern LLMs and two traditional benchmarks. We apply these methods with four language models: GPT-4~\citep{openai}, Claude 3 Sonnet, LLaMA-3-Chat (70B)~\citep{llama3modelcard}, and LLaMA-2-Chat (70B)~\citep{llama2}. To provide a gold standard for assessing the effectiveness of these methods, we create an oracle using LLaMA-2 (70B) by intentionally contaminating the model with varying portions of the six challenging benchmarks in the format of instruction fine-tuning with answer augmentation. This setup allows us to observe the performance of the five detection methods given the known contamination status. By these setups, we seek to answer the following research questions: \textit{i.} Are the latest state-of-the-art LLMs, which consistently achieve higher performance, contaminated with these challenging benchmarks? What do the detection methods indicate? \textit{ii.} Can these methods detect contamination that occurred during instruction fine-tuning with data variations instead of the original format in pretraining? \textit{iii.} Do different ``well-accepted'' detection methods corroborate each other's findings for a given dataset? Do they yield inconsistent results? 

Our experimental results and analysis reveal several critical findings about current data contamination detection in LLMs: \textit{First}, all existing methods have limitations in their underlying assumptions or practical applications. 
\textit{Second}, while some metrics suggest possible contamination in traditional benchmarks, we observe no consistent agreement for methods contamination in newer, more challenging benchmarks. \textit{Third}, all methods struggle to robustly reflect our oracle contamination created by instruction fine-tuning with answer augmentation. This finding highlights an urgent need for this research direction. \textit{Finally}, we observe a surprising lack of agreement between different detection methods, suggesting that these well-accepted approaches cannot be simultaneously valid. This disagreement casts doubt on the reliability of current contamination detection techniques and highlights the critical need for more robust, consistent, and comprehensive approaches.


\begin{figure*}[!thb]
    \centering
    \includegraphics[width=\textwidth]{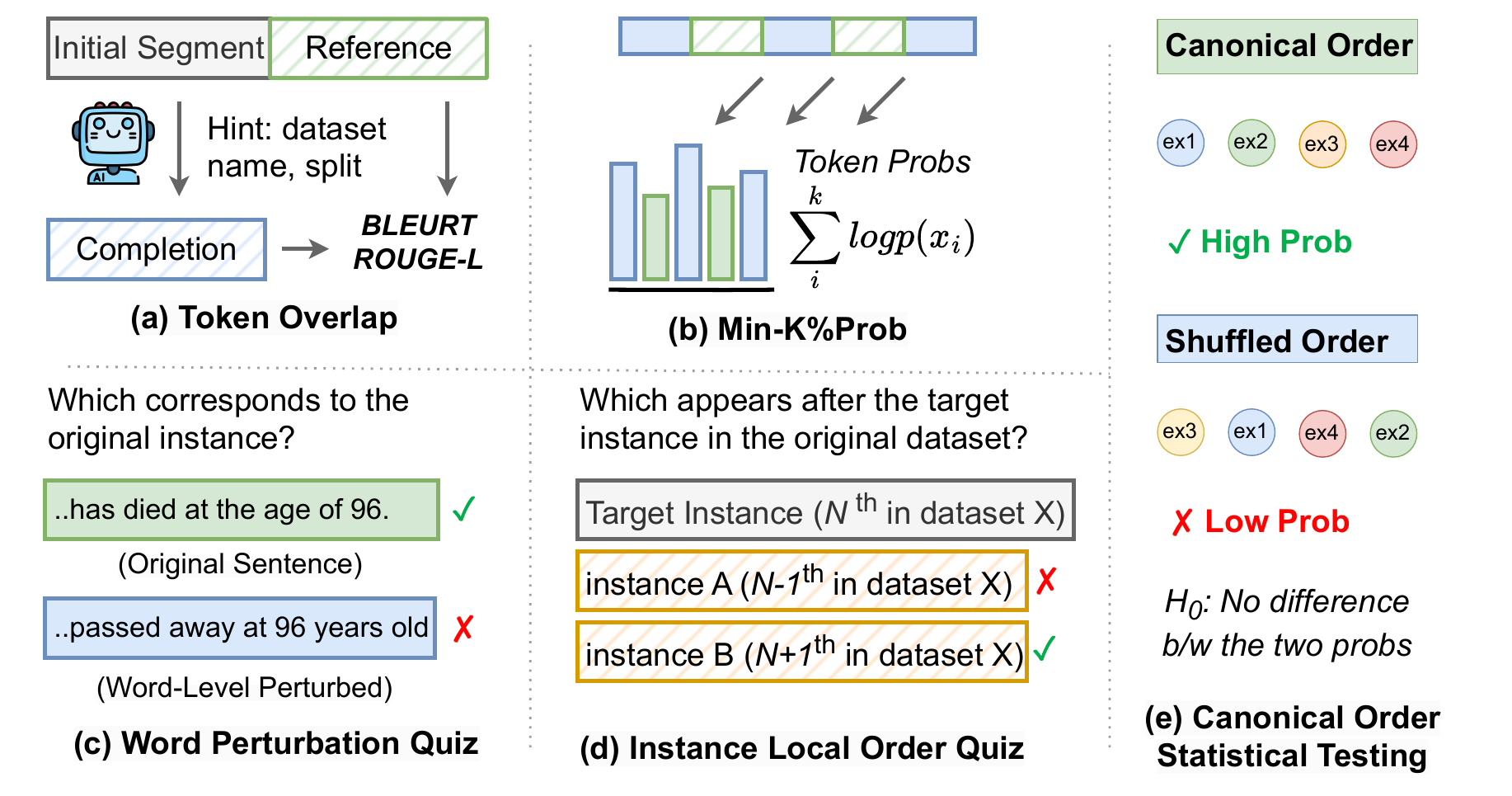} 
    \caption{Overview of five methods evaluated for detecting data contamination in large language models (LLMs). \textbf{(a) Token Completion Overlap Score:} Evaluates LLM contamination by prompting the model with partial text to see if the completion matches a reference instance. \textbf{(b) Min-K\% Probability:} Determines potential contamination by assessing the average log-likelihood of the k\% least probable tokens in a text. \textbf{(c) Word Perturbation Quiz:} Detects contamination by checking if the LLM can distinguish between a word-level perturbed instance and the original. \textbf{(d) Instance Local Order Quiz:} A prompt-based method we developed, assessing whether the LLM can identify the correct subsequent example in a dataset sequence. \textbf{(e) Canonical Order Statistical Testing:} Tests if the LLM shows a preference for canonical order over random shuffling, indicating potential contamination.}
    \label{fig:pipeline}
\end{figure*}

\section{Related Work} 


Detecting training data through probability inference and reconstruction has long been a well-established approach~\cite{MIA,gpt2-reconstruct}.
Recently, the challenge of data contamination in large language models has garnered significant attention due to its potential to skew model evaluation and misrepresent true performance. \citet{sainz2023nlp} highlighted the risks of contamination, particularly emphasizing that while test set contamination invalidates benchmarks, contamination in training and validation sets is less concerning unless zero or few-shot learning claims are made. Contamination is most likely during the pre-training phase, where massive text corpora are scraped with minimal curation. Additionally, \citet{balloccu2024leak} found that 42\% of papers evaluating models such as GPT-3.5 and GPT-4 contained leaked data, affecting millions of instances, further underscoring the widespread impact of contamination.

Several methods have been proposed to detect data contamination, focusing on either log probability analysis or prompting-based techniques. Log probability-based methods, such as those developed by \citet{contaminat-baseline-canonical-p-value-ICLR24-oral} and \citet{contaminat-baseline-min-k-prob-AUC-ICLR24}, assess the likelihood of data being present in a model's training set. In contrast, prompting-based approaches by \citet{contaminat-baseline-completion-overlap-ICLR24-spotlight} and \citet{contaminat-baseline-perturb-quiz} directly query the model to detect contamination. However, these methods have primarily been tested on traditional benchmarks and early-stage models, leaving a gap in understanding their effectiveness on more recent, advanced LLMs and complex datasets.


Besides, recent work by \citet{yao-etal-2024-data} introduces a generalization-based approach, highlighting how cross-lingual contamination can bypass traditional detection methods by inflating performance when benchmarks are translated into other languages. This study shows that models contaminated in this way fail to generalize effectively when answer options are manipulated, offering a novel angle for detecting deeply concealed contamination. 

Moreover, much of the existing research assumes contamination occurs during pre-training, overlooking the potential impact of instruction fine-tuning, which is increasingly used to enhance LLM capabilities. This stage can introduce variations in the data that are not adequately tested by current methods. Our work seeks to address these gaps by evaluating a broader range of detection techniques across diverse benchmarks and models, with particular attention to more challenging datasets and the instruction fine-tuning phase.

\section{Benchmark Datasets}

There exists a significant disparity between the benchmarks commonly used in data contamination research and those employed to evaluate state-of-the-art (SOTA) LLM capabilities. Table~\ref{tab:dataset-mismatch} in the Appendix illustrates this mismatch, showing that while many contamination detection methods are validated using traditional benchmarks, modern LLM evaluations focus on more challenging tasks, such as mathematical reasoning and code generation. This observation raises intriguing questions: Could the performance gains of SOTA LLMs on these newer, more challenging benchmarks be attributed, in part, to data contamination? What insights do SOTA data contamination detection methods provide when applied to these essential modern benchmarks?

To address this, we selected six challenging benchmarks commonly used in evaluating SOTA LLMs, complemented by two traditional benchmarks frequently featured in data contamination studies. We aim to ensure broad coverage of relevant task domains while allowing for comparison with previous contamination detection work. The selected benchmarks are as follows:

\vspace{0.02in}
\noindent \textbf{$\bullet$ GSM8K}~\citep{GSM8K} contains linguistically diverse grade school-level math questions with moderate difficulties. 

\vspace{0.02in}
\noindent \textbf{$\bullet$ MMLU}~\citep{MMLU} contains multiple choice questions across multiple domains.

\vspace{0.02in}
\noindent \textbf{$\bullet$ BIG-Bench-Hard (BBH)}~\citep{BBH} contains multitask questions believed to be beyond the capabilities of LLMs at the time of release.

\vspace{0.02in}
\noindent \textbf{$\bullet$ ARC-Challenge}~\citep{arc} contains questions from the ARC dataset that were answered incorrectly by both a retrieval-based algorithm and a word co-occurrence algorithm. 

\vspace{0.02in}
\noindent \textbf{$\bullet$ DROP}~\citep{drop} a new reading comprehension benchmark requiring discrete reasoning over paragraphs. 

\vspace{0.02in}
\noindent \textbf{$\bullet$ HumanEval}~\citep{humaneval} is a challenging becnhamrk for the coding domain. 

\vspace{0.02in}
\noindent \textbf{$\bullet$ AGNews}~\citep{agnews} contains text classification questions drawn from over 1 million news articles

\vspace{0.02in}
\noindent \textbf{$\bullet$ IMDB}~\citep{imdb} is a binary sentiment classification datasets from movie reviews.

\section{Evaluated Methods and Limitations}

We examine five distinct approaches to detecting data contamination in LLMs, including three state-of-the-art techniques from ICLR (2023-2024) and two exploratory prompt-based approaches. Two approaches are based on sequence probabilities and require access to model parameters. Figure \ref{fig:pipeline} illustrates the visual overview of each approach. For each method, we also note the {limitations} we identified during our examination. The overview of these methods are illustrated in Figure~\ref{fig:pipeline}.

\vspace{0.05in}
\noindent $\bullet$ \textbf{Min-K\% Prob}~\citep{contaminat-baseline-min-k-prob-AUC-ICLR24} assesses whether a text was in an LLM's pre-training data by calculating the average log-likelihood of the k\% lowest-probability tokens, with a high result suggesting the text's presence in the training data. 

\noindent \textbf{Limitations:} (1) The authors report AUC based on the proposed WiKiMIA dataset, in which they regarded data events before the model release as contaminated data. Such a strong assumption on the ground truth may require more justification. (2) They did not provide the threshold to determine the value of min-K\%-prob in the paper since they claim they can use AUC; however, in real-world settings, we do not always have the oracle to determine AUC - instead, we need a metric for determining whether arbitrary datasets are contaminated. (3) The code is not available. 

\vspace{0.05in}
\noindent $\bullet$ \textbf{Canonical Order Statistical Testing}~\citep{contaminat-baseline-canonical-p-value-ICLR24-oral} identifies contamination in a pre-training dataset by checking if the model shows a preference for the canonical order of examples over random shuffling. This preference is tested by comparing their log probabilities, with results aggregated across datasets to ensure a low false positive rate.

\noindent \textbf{Limitations:} When an individual data example's length is long, the combination of sample/shards/permutations in the setup can be costly.

\vspace{0.05in}
\noindent $\bullet$ \textbf{Token Completion Overlap Score}~\citep{contaminat-baseline-completion-overlap-ICLR24-spotlight} detects contamination by prompting the LLM with a dataset name, partition type, and a random initial segment of a reference instance. If the LLM's output closely matches the latter part of the reference, the instance is flagged as contaminated.

\noindent \textbf{Limitations:} (1) Part of the evaluation is by GPT-4, prompting GPT-4 to determine a ``near match'' can be ambiguous and subject to biases. (2) It is unclear how different parts of the original data points can affect the completion and, thus, the ROUGE score and p-value.

\vspace{0.05in}
\noindent $\bullet$ \textbf{Word Perturbation Quiz}~\citep{contaminat-baseline-perturb-quiz} detects data contamination by presenting an LLM with a multiple-choice quiz, where the options include word-level perturbed versions of a dataset instance and the original. The LLM's tendency to select the original instance indicates potential contamination from its pre-training.

\noindent \textbf{Limitations:} While perturbed answers may retain their semantic meaning, they often lack the natural fluency of the original text. This discrepancy in linguistic nuance can inadvertently provide cues to the model, making it easier to identify the unperturbed, ground truth answer. Moreover, the original perturbation prompts do not safeguard proper nouns and numerical values from alteration. Given that the labels remain unperturbed, these distinctive elements can serve as additional indicators for the model to differentiate between original and perturbed content. Consequently, the model's ability to select the correct answer may stem from recognizing these linguistic and contextual inconsistencies rather than from accurate memorization or contamination, potentially leading to overestimating contamination levels.

\vspace{0.05in}
\noindent $\bullet$ \textbf{Local Order Quiz (Ours)} In this work, we are also interested in exploring whether prompt-based approaches can detect data contamination in LLMs. These approaches could offer the advantage of being applied to both closed-source and open-source models. Unlike token perturbation methods, which may entangle an LLM's recognition of original data points with its sensitivity to perturbations, we propose an approach focusing on the model's ability to identify the original order of dataset examples. Specifically, we randomly sample a target example $t$ from dataset $D$ and provide $N$ other examples, one of which appeared immediately after $t$ in the original dataset. We then prompt the LLM to identify this subsequent example. While pretraining typically involves randomized batches, local order can be preserved to some extent. If the LLM can accurately identify the correct subsequent example, especially in cases with no inherent information suggesting the order (e.g., similar categories or content), this could indicate potential data contamination. \textbf{Limitations:} We acknowledge the challenging nature of this task, as it requires the LLM not only to have been exposed to the data points in their original order but also to understand their semantics and retrieve this information accurately. 

\section{Experiments}

\begin{table*}[h!]
    \centering
    \begin{adjustbox}{width=0.92\textwidth}
    \begin{tabular}{@{}lccccccc@{}}
        \toprule
        \textbf{Model} & \textbf{Datasets} & \textbf{Split} & \textbf{WPQ } & \textbf{Ours} & \textbf{Token Overlap} & \textbf{Min-K\%} & \textbf{Canonical Order} \\
          &  &  & \small{\textit{(accuracy)}} & \small{\textit{(accuracy)}} & \small{\textit{exact/near/p-value}} & \small{\textit{$\text{Mean}_{\text{var}}$}} & \small{\textit{p-value}} \\
        \midrule
        GPT-4 & MMLU & test & 0.68 & 0.24 & 0/0/0.23 & - & - \\
        & BBH & - & 0.65 & 0.28 & 0/0/0.10 & - & - \\
        & ARC-Challenge & train & 0.67 & 0.29 & 1/0/0.47 & - & - \\
        & ARC-Challenge & test & 0.76 & 0.28 & 0/0/0.03 & - & - \\
        & DROP & train & 0.64 & 0.20 & 0/0/0.62 & - & - \\
        & DROP & test & 0.57 & 0.21 & 0/0/0.99 & - & - \\
        & HumanEval & test & 0.83 & 0.36 & 0/0/0.21 & - & - \\
        & GSM8K & train & 0.72 & 0.36 & 0/1/0.21 & - & - \\
        & GSM8K & test & 0.74 & 0.22 & 0/0/0.65 & - & - \\
        & AG News & train & 0.72 & 0.42 & 0/0/0.10 & - & - \\
        & AG News & test & 0.81 & 0.35 & 0/0/0.18 & - & - \\
        & IMDB & train & 0.79 & 0.81 & 0/0/0.72 & - & - \\
        & IMDB & test & 0.76 & 0.83 & 0/0/0.01 & - & - \\
        \midrule
        Claude 3 & MMLU & test & 0.29 & 0.21 & 0/0/1.00 & - & - \\
        & BBH & - & 0.48 & 0.28 & 0/0/0.98 & - & - \\
        & ARC-Challenge & train & 0.48 & 0.30 & 0/0/1.00 & - & - \\
        & ARC-Challenge & test & 0.48 & 0.25 & 0/0/0.97 & - & - \\
        & DROP & train & 0.56 & 0.28 & 0/0/0.96 & - & - \\
        & DROP & test & 0.53 & 0.25 & 0/0/0.24 & - & - \\
        & HumanEval & test & 0.90 & 0.24 & 0/0/0.06 & - & - \\
        & GSM8K & train & 0.90 & 0.26 & 0/0/0.96 & - & - \\
        & GSM8K & test & 0.91 & 0.16 & 0/0/0.53 & - & - \\
        & AG News & train & 0.79 & 0.27 & 0/0/0.61 & - & - \\
        & AG News & test & 0.74 & 0.28 & 0/0/0.82 & - & - \\
        & IMDB & train & 0.64 & 0.65 & 0/0/0.08 & - & - \\
        & IMDB & test & 0.60 & 0.65 & 0/0/0.36 & - & - \\
        \midrule
        LLaMA 3 & MMLU & test & 0.47 & 0.23 & 0/0/0.20 & 8.83$_{2.5}$ & 0.69 \\
        & BBH & - & 0.48 & 0.20 & 0/0/0.38 & 10.14$_{6.5}$ & 0.11 \\
        & ARC-Challenge & train & 0.56 & 0.21 & 0/0/0.80 & 9.18$_{2.0}$ & 0.92 \\
        & ARC-Challenge & test & 0.41 & 0.23 & 0/0/0.30 & 9.08$_{2.0}$ & 0.97 \\
        & DROP & train & 0.48 & 0.24 & 0/0/0.19 & 6.92$_{0.7}$ & 0.20 \\
        & DROP & test & 0.38 & 0.32 & 0/0/1.00 & 6.53$_{1.0}$ & 0.85 \\
        & HumanEval & test & 0.85 & 0.31 & 0/0/0.40 & 7.42$_{1.6}$ & 0.59 \\
        & GSM8K & train & 0.68 & 0.29 & 0/0/0.81 & 7.67$_{1.3}$ & 0.88 \\
        & GSM8K & test & 0.70 & 0.25 & 0/0/0.64 & 7.88$_{1.7}$ & 0.25 \\
        & AG News & train & 0.54 & 0.31 & 0/0/0.33 & 11.55$_{1.5}$ & 0.98 \\
        & AG News & test & 0.69 & 0.33 & 0/0/0.60 & 11.58$_{1.7}$ & 0.12 \\
        & IMDB & train & 0.38 & 0.77 & 0/0/0.07 & 9.33$_{1.2}$ & 0.60 \\
        & IMDB & test & 0.41 & 0.86 & 0/0/0.01 & 9.13$_{2.5}$ & 0.21 \\
        \bottomrule
    \end{tabular}
    \end{adjustbox}
    \caption{We evaluated contamination in GPT-4, Claude 3 Sonnet, and LLaMA-3-70b using five methods across eight datasets. The Token Perturbation Quiz (WPQ) assessed model accuracy in selecting correct examples from semantically similar options (n=100). Our method measured accuracy in predicting subsequent dataset examples (n=100). Token overlap analysis compared guided and unguided completions, reporting exact matches, near matches, and p-value for ROUGE-L score differences between guided and general instructions (n=10). Min-K\% calculations yielded mean log probabilities and standard deviations (k=20, n=100). Canonical order testing compared likelihoods between original and shuffled dataset orders (10 shards, 25 permutations).}
    \label{tab:main_results}
\end{table*}

\begin{table*}[h!]
    \centering
    \small
    \begin{tabular}{@{}lcccccccc@{}}
        \toprule
        \textbf{Datasets} & \textbf{Split} & \textbf{Contamination} & \textbf{WPQ } & \textbf{Ours} & \textbf{Token Overlap} & \textbf{Min-K\%} & \textbf{Canonical Order} \\
           & & (\%) & \tiny{\textit{(accuracy)}} & \tiny{\textit{(accuracy)}} & \tiny{\textit{exact/near/p-value}} & \tiny{\textit{$\text{Mean}_{\text{var}}$}} & \tiny{\textit{p-value}} \\
        \midrule
        MMLU & test & - & 0.29 & 0.26 & 0/0/0.090 & 7.40$_{2.0}$ & 0.99 \\
        BBH & - & - & 0.33 & 0.23 & 0/0/0.053 & 8.40$_{3.8}$ & 0.00 \\
        ARC-Challenge & train & - & 0.27 & 0.27 & 0/0/0.034 & 7.72$_{1.8}$ & 0.15 \\
        ARC-Challenge & test & - & 0.25 & 0.21 & 0/0/0.154 & 7.72$_{2.2}$ & 0.80 \\
        DROP & train & - & 0.29 & 0.23 & 0/0/0.697 & 6.04$_{0.7}$ & 0.07 \\
        DROP & test & - & 0.38 & 0.31 & 0/0/0.453 & 5.50$_{1.1}$ & 0.16 \\
        HumanEval & test & - & 0.75 & 0.28 & 0/4/0.039 & 6.88$_{1.1}$ & 0.07 \\
        GSM8K & train & - & 0.60 & 0.27 & 0/1/0.229 & 7.87$_{1.2}$ & 0.61 \\
        GSM8K & test & - & 0.62 & 0.21 & 0/0/0.311 & 7.88$_{1.1}$ & 0.41 \\
        \midrule
        MMLU & test & 100 & 0.33 & 0.24 & 0/0/0.477 & 5.46$_{1.9}$ & 0.81 \\
        BBH & test  & 100 & 0.38 & 0.21 & 0/0/0.833 & 4.47$_{2.8}$ & 0.10 \\
        ARC-Challenge & train & 50 & 0.29 & 0.22 & 0/0/0.211 & 5.94$_{1.1}$ & 0.33 \\
        ARC-Challenge & test & 0 & 0.31 & 0.36 & 0/0/0.035 & 5.99$_{1.7}$ & 0.57 \\
        DROP & train & 50 & 0.29 & 0.31 & 0/0/0.767 & 4.28$_{1.0}$ & 0.01 \\
        DROP & test & 0 & 0.33 & 0.24 & 0/0/0.646 & 4.07$_{1.1}$ & 0.30 \\
        HumanEval & test & 25 & 0.76 & 0.14 & 0/0/0.021 & 5.16$_{0.9}$ & 0.36 \\
        GSM8K & train & 25 & 0.37 & 0.22 & 0/0/0.473 & 6.12$_{0.9}$ & 0.28 \\
        GSM8K & test & 0 & 0.57 & 0.30 & 0/0/0.093 & 6.23$_{0.9}$ & 0.07 \\
        \bottomrule
    \end{tabular}
    \caption{The oracle contamination detection results with LLaMA-2-70b, before (top) and after (bottom) instruction fine-tuning with the corresponding proportion of the datasets.}
    \label{tab:oracle_results}
\end{table*}

\begin{table*}[h]
    \centering
    \begin{adjustbox}{width=0.75\textwidth}
    \captionsetup{font=small,labelfont=bf}
    \begin{tabular}{lccccc}
        \toprule
        & \textbf{Token Overlap} & \textbf{WPQ} & \textbf{Ours} & \textbf{Min-K\%} & \textbf{Canonical Order} \\
        \midrule
        \rowcolor{gray!20} \textbf{Token Overlap} & 1.000 & -0.067 & -0.009 & -0.198 & 0.096 \\
        \textbf{WPQ} & -0.067 & 1.000 & -0.014 & 0.320 & 0.076 \\
        \rowcolor{gray!20} \textbf{Ours} & -0.009 & -0.014 & 1.000 & -0.063 & -0.187 \\
        \textbf{Min-K\%} & -0.198 & 0.320 & -0.063 & 1.000 & 0.040 \\
        \rowcolor{gray!20} \textbf{Canonical Order} & 0.096 & 0.076 & -0.187 & 0.040 & 1.000 \\
        \bottomrule
    \end{tabular}
    \end{adjustbox}
    \caption{Spearman correlation between the 5 methods of data contamination studied in this work. We calculate correlation values across all results obtained by our 4 vanilla models (GPT-4, Claude 3 Sonnet, LLaMA-3-70b, LLaMA-2-70b) as well as our oracle model.}
    \label{tab:metric_correlations}
\end{table*}

In this section, we describe our experimental settings for each method and the oracle, and the experiment results.

\subsection{Implementation Details} For all detection methods, we follow their official implementations when available. For the Oracle setup, we instructed fine-tuned LLaMA-2-70b-chat using the original answers of the examples replaced by chain-of-thought reasoning. Specifically: 

\paragraph{Min-K\% Prob} We used $k=20$ as recommended in the paper. The authors used their proposed WikiMIA dataset as ground truth and reported AUC in the original paper. However, the threshold for determining contamination is not provided, and in real-world settings, the ground truth is unavailable. Thus, we used the mean and standard deviation for each split directly for the contamination indicator, as the authors claim Min-K\% Prob is most informative compared with other probability-based metrics.

\paragraph{Canonical Order Statistical Testing} Due to resource constraints, we adapted the method to use 100 instances with 10 shards and 25 permutations. A shard is a partition of the overall dataset into an equal sized portion. Examples within each shard were concatenated with \symbol{92}n. All the datasets were processed so that only the question and answer were included for each instance. 

\paragraph{Token Completion Overlap Score} We used the exact implementation without changes, including only ten instances. GPT-4 was employed to determine near/exact matches, with one exact match and two near matches used as the threshold for contamination detection. The temperature was set to 0 for inference. The exact ICL prompt used for GPT-4 evaluation is shown in Figure~\ref{fig:icl_prompt} in the Appendix.

\paragraph{Word Perturbation Quiz} LLaMA-3-70b-chat was used for perturbation across all datasets. The perturbation prompt was adjusted for different dataset formats, with care to prevent altering proper nouns. For perturbation, temperature and top p were set to 0.9, while for inference, temperature was 0. Dataset-specific adjustments included: Humaneval: Only docstrings were perturbed. DROP: Full passage and question were perturbed. MMLU/AGNews/IMDB/ARC-Challenge/GSM8K: Answer choices were removed before perturbation to avoid confusion in the quiz.

\paragraph{Local Order Quiz (Ours)} We randomly selected options for each instance, ensuring options from the same category for datasets like MMLU and BBH. The prompt included a dataset description, name, data split, and options. We used the number of options of 4 as a hyperparameter, meaning a random guess would achieve about 25\% accuracy. All datasets maintained their original format as initially processed.

\paragraph{Oracle Setup} To investigate the challenges of detecting data contamination during the fine-tuning stage, we developed an oracle setup that mimics real-world instruction fine-tuning scenarios. We hypothesize that contamination occurring during fine-tuning is significantly more difficult to detect compared to pretraining contamination for several key reasons: \textbf{(1) Exposure Frequency}: During fine-tuning, the model typically encounters data for only a few epochs (usually 1-3), whereas in pretraining, data chunks are often seen repeatedly due to sliding window approaches ([starting position: starting position + window size]). \textbf{(2) Data Modification}: Modern fine-tuning techniques, particularly instruction fine-tuning, often incorporate more complex answers, such as chain-of-thought reasoning, to enhance the LLM's analytical capabilities. This process modifies the original data, preserving the questions while replacing answers with elaborate solutions. To simulate these conditions, we fine-tuned LLaMA-2-70B-chat using full fine-tuning with varying proportions of the six newer benchmarks in our study. Our goal was to observe how different contamination detection metrics respond to varying levels of data exposure during fine-tuning. For each data point, we replaced the original answer with a chain-of-thought solution generated by LLaMA-2-70B, mimicking the data augmentation often used in instruction fine-tuning. To maintain some semblance of the original data structure, we packaged four examples as one training instance, preserving the local order as it appears in the original dataset. This allows to test whether detection methods can identify contamination when local order information is partially retained. The fine-tuning process used a learning rate of 8e-6 and ran for three epochs, aligning with typical fine-tuning practices. By varying the proportion of benchmark data used in this process, we aimed to create a controlled environment where the degree of contamination is known, allowing us to evaluate the sensitivity and reliability of various detection methods in identifying fine-tuning stage contamination.

The prompt templates used in the paper are available in the Appendix.

\subsection{Main Results} 
Table~\ref{tab:main_results} shows the results of the five contamination detection methods on eight benchmarks with GPT-4 (\texttt{gpt-4-0613}), Claude-3 Haiku, and LLaMA-3-70b. Results for the Min-K\% and Canonical Order methods are unavailable for GPT-4 and Claude-3, as these approaches require access to model parameters. We can observe that: (1) Perturbation accuracy is unusually high across most benchmarks for all three models. This consistent pattern suggests potential inflation of results, possibly due to the models recognizing the perturbations themselves rather than indicating contamination or memorization. (2) Both our proposed method and the token overlap approach provide significant evidence that the IMDB dataset may be contaminated for all three models. (3) The token overlap method yields conflicting results depending on the specific metric. Based on p-values, it suggests contamination in HumanEval for Claude-3 and in BBH for GPT-4. However, no contamination is detected for any of the newer benchmarks when considering exact matches and near matches. (4) For LLaMA-3, both Min-K\% and Canonical Order P-value show possible contamination on BBH. However, in general, there is no clear agreement among the methods regarding contamination in the newer benchmarks. Notably, the GPT-4 report indicates that DROP and HumanEval are approximately 21-25\% contaminated. However, our results show that none of the methods detected contamination in DROP, while only our method partially detected contamination in HumanEval (accuracy = 0.36).

Table~\ref{tab:oracle_results} presents the oracle results, comparing the metric values of the five detection methods before (upper half) and after (lower half) instruction fine-tuning. The Portion column indicates the percentage of data used for fine-tuning, ranging from 0\% (no contamination) to 100\% (entire dataset split used). Ideally, a robust detection method should (1) consistently reflect the portion of contamination through its metric values and (2) not be influenced by the training set contamination when the test set remains uncontaminated. For instance, probability-based approaches might be susceptible to false positives in test set contamination detection when the training set is contaminated due to the similarity in distribution between training and testing data. Therefore, post-fine-tuning results should show increased accuracy or mean probability values and decreased p-values for non-zero contamination portions in an ideal scenario. However, our observations reveal significant challenges in detecting fine-tuning contamination: (1) None of the metrics demonstrate a consistent trend that aligns with the varying portions of data contamination. This lack of correlation suggests that current methods may not be sensitive enough to detect or quantify the degree of contamination introduced during such a fine-tuning paradigm. (2) Surprisingly, all Min-K\% probability values decrease after fine-tuning. This counterintuitive result is particularly noteworthy given that the original questions are preserved in the fine-tuning process, even though we used instruction fine-tuning with chain-of-thought solutions. These findings highlight the complexity of detecting contamination in instruction-tuned models and suggest that existing methods, originally designed for pretraining contamination detection, may not be directly applicable or reliable in fine-tuning scenarios. 

Table~\ref{tab:metric_correlations} shows the Spearman rank correlation between the metric results of the five detection methods based on all experiments conducted. While we acknowledge slight statistical liberty in comparing rank correlations between p-values and other metrics, this analysis provides insights into the relationships between different contamination detection approaches.

First, we observe no strong correlations or agreement between these metrics across the various LLMs and datasets tested. This lack of consensus is particularly concerning, as it suggests that different methods may yield contradictory conclusions about the presence or extent of contamination in a given model-dataset pair. We do, however, note some weak correlations: (1) Between the Min-K\% probability value and word perturbation quiz accuracy, (2) Between Min-K\% probability and token overlap p-values, and (3) Between canonical order p-values and our proposed method. The weak correlation between our method and the canonical order statistical testing suggests the potential for prompting information about local order as a proxy for canonical statistical testing, especially in gauging data contamination in closed-source LLMs. 

However, the lack of strong agreement between methods raises a critical concern: these well-established approaches cannot all be simultaneously correct in their contamination assessments, which poses a significant challenge to contamination detection in LLMs. This observation calls for further investigation into the more robust and consistent contamination detection methods for distinct scenarios.

\section{Conclusions and Future Work}

This paper evaluates five distinct data contamination detection methods across eight benchmarks and four state-of-the-art LLMs, including an oracle setup to mimic instruction fine-tuning contamination. Our study reveals significant challenges in current methods, with inconsistent results across different benchmarks and models. Detecting contamination introduced during instruction fine-tuning proved especially difficult, and the weak correlations between different detection methods raise concerns about their collective reliability. While prompt-based methods show certain evidence of detection abilities, they are very limited at this stage. These findings underscore the complexities in accurately quantifying data contamination in LLMs and highlight the urgent need for more robust, unified detection frameworks. As LLMs continue to advance, developing reliable contamination detection techniques remains crucial for ensuring the integrity and trustworthiness of AI systems.

\section*{Limitations}
  While we discovered significant disagreement between different contamination detection methods, especially compared to oracle tests, our investigation did not uncover the underlying reasons for these discrepancies. Additionally, our proposed prompt-based contamination detection method showed limited effectiveness, highlighting the inherent challenges in using prompting for this task.

\section*{Ethics Statement}
This study on data contamination detection in LLMs has ethical implications on the importance of transparency in AI development and the potential risks of overestimating model capabilities based on potentially contaminated evaluations. We emphasize that our findings on the limitations of current detection methods call for caution in making definitive claims about data contamination and highlight the urgent need for more robust and general detection methods. 

\DeclareRobustCommand{\disambiguate}[3]{#3}
\bibliography{custom}
\clearpage

\appendix

\setcounter{table}{0}
\renewcommand{\thetable}{A\arabic{table}}

\section{Prompts}
\label{sec:prompts}
\subsection{Word Perturbation Quiz}
As part of the Word Perturbation Quiz, there are two components namely the perturbation and the quiz components. In the perturbation prompt, the model is prompted to create a four choice quiz by only making word level perturbations. Several guidelines are also included in the prompt to guide the model on what words are allowed to perturbed and what formatting components must remain the same. In the standard quiz component, the model is given the 3 perturbed options and the original data instance in a random order and prompted to select the correct data instance that showed up in the given dataset and split. Both of these prompts closely follow the prompts in ~\citep{contaminat-baseline-perturb-quiz}.

\phantomsection
\label{appendix:dca_prompt1}
\noindent
{\small
\begin{minipage}[!htb]{\linewidth}
\phantomsection
    \begin{tcolorbox}[colback=white!95!gray,colframe=gray!50!black,rounded corners,label={box:question-gen-prompt}, title={Perturbation Prompt.}]
Instruction: Your task is to create a four-choice quiz by replacing the words in the provided "Input Text" with their contextually relevant synonyms. The meaning and overall structure of the four options must exactly match every detail and the structure in the Input Text. You must not include the provided Input Text as an option. Each option in the four-question quiz you generate must include both the underlying text and answer choices in the Input Text but with word-level perturbations.
You must make sure that:\\
\\
(1) You generate distinct options based on the provided Input Text;\\
(2) The only difference between options is word-level perturbations.\\ 
(3) Each Option must still include the main part of the text in the Input Text with word level perturbations\\
(4) Each option must still include all answer choices present in the Input Text with no changes\\
(5) No numbers or proper nouns are changed.\\
(6) Options are ordered;\\
(7) There is not any extra explanation;\\
(8) You comply with the following "Format" to generate options\\
(9) You comply with every specific symbol and letter detail in the given Input Text; and \\
(10) All options retain the exact label from the Input Text, if there is one.\\
---\\
Input Text:\\
{instance}\\
---\\
Format:\\
Option 1 - \\

Option 2 - \\

Option 3 - \\

Option 4 - \\

    \end{tcolorbox}
\end{minipage}}


\phantomsection
\label{appendix:dca_prompt2}
\noindent
{\small
\begin{minipage}[thb]{\linewidth}
\phantomsection
    \begin{tcolorbox}[colback=white!95!gray,colframe=gray!50!black,rounded corners,label={box:question-gen-prompt}, title={Standard Quiz Prompt.}]
Instruction: You are provided with a five-choice quiz. Your task is to correctly select the option that exactly corresponds to an instance from the [split] split of the [dataset] dataset.\\
\\
When selecting the option, you must ensure that you follow the following rules:\\
(1) You must ensure that you only generate a single option letter as your answer.\\
(2) If you do not know the dataset or the correct answer, you must select option "(E) None of the provided options."\\
(3) You must output the answer in your final sentence like "The correct answer is ..."\\
\\
Hint: While all the following options seem similar in terms of meaning, there is only one option that reflects the exact wording with respect to the original instance.\\
---\\
(A) {choices["(A)"]}\\
(B) {choices["(B)"]}\\
(C) {choices["(C)"]}\\
(D) {choices["(D)"]}\\
(E) {choices["(E)"]}\\
---\\
Answer:

    \end{tcolorbox}
\end{minipage}}

\subsection{Local Order Quiz}
In the local order quiz, the model is given a description of the dataset, a specific example from the dataset, and options for the next occurring example in order in the dataset. The model is then queried to choose which of the given options is the next occurring example in the dataset in order.  The description of the the dataset is pulled from the GitHub page or Hugging Face page of the dataset if once exists. The rationale here is that while web scraping, if a dataset was contaminated then the dataset description on the GitHub/Hugging Face would have been contaminated as well. If a description foes not exist on the Github or Hugging Face then the description from the paper related to the dataset is used.

\phantomsection
\label{appendix:dca_prompt2}
\noindent
{\small
\begin{minipage}[t]{\linewidth}
\phantomsection
    \begin{tcolorbox}[colback=white!95!gray,colframe=gray!50!black,rounded corners,label={box:question-gen-prompt}, title={Local Order Quiz Prompt.}]
[Description of dataset]\\
\\
Given the target data example in the [split] of the [dataset name] dataset, Which of the following examples was next to it in the original order of the dataset? Exactly one of the choices must be selected and you need to output the answer in your final sentence like "The answer is ..."\\
\\
Target example: [tar example]\\
\\
Options: [options]

    \end{tcolorbox}
\end{minipage}}

\newpage
\clearpage
\phantomsection
\noindent
{\small
\begin{minipage}[t]{\textwidth}
\phantomsection
    \begin{tcolorbox}[colback=white!95!gray,colframe=gray!50!black,rounded corners,label={box:rubrics}, title={GPT-4 ICL prompt used for Token Overlap}]
\textbf{INSTRUCTION:}\\
You are provided with a reference text and a candidate text.
The candidate text is a generated replica of the reference text.
Your task is to determine if the candidate text is an exact or near-exact match of the reference text.\\
Near-exact match candidates must PRECISELY match the reference candidates in terms of sentence structure, overlap, and contextual similarity.
Respond only with 'Yes' or 'No'.\\
\\
\textbf{Example 1:}\\
\\
REFERENCE TEXT:\\
The cat waited at the top.\\
\\
CANDIDATE TEXT:\\
The cat waited at the top.\\
\\
ANSWER: Yes (exact match)\\
\\
\textbf{Example 2:}\\
\\
REFERENCE TEXT:\\
icy surface of Jupiter's largest moon, Ganymede. These irregular masses may be rock formations, supported by Ganymede's icy shell for billions of years.\\
\\
CANDIDATE TEXT:\\
icy surface of Jupiter's largest moon, Ganymede. These irregular masses may be rock formations, supported by Ganymede's icy shell for billions of years. This discovery supports the theory that Ganymede has a subsurface ocean. Scientists used gravity data from NASA's Galileo spacecraft to create a geophysical model of the interior of Ganymede.\\
\\
ANSWER: Yes (near-exact match)\\
\\
\textbf{Example 3:}\\
\\
REFERENCE TEXT:\\
50th Anniversary of Normandy Landings lasts a year.\\
\\
CANDIDATE TEXT:\\
The 50th anniversary celebration of the first Normandy landing will last a year.\\
\\
ANSWER: Yes (near-exact match)\\
\\
\textbf{Example 4:}\\
\\
REFERENCE TEXT:\\
Microsoft's Hotmail has raised its storage capacity to 250MB.\\
\\
CANDIDATE TEXT:\\
Microsoft has increased the storage capacity of its Hotmail e-mail service to 250MB.\\

ANSWER: Yes (near-exact match)\\
\\
\textbf{Example 5:}\\
\\
REFERENCE TEXT:\\
\{reference\_text\}\\
\\
CANDIDATE TEXT:\\
\{candidate\_text\}\\
\\
ANSWER:
    \end{tcolorbox}
\label{fig:icl_prompt}
\end{minipage}}
\clearpage
\begin{table*}[tbh]
\centering
\resizebox{\textwidth}{!}{%
\begin{tabular}{@{}ccccccccc@{}}
\toprule
\textbf{Dataset} & \textbf{LLaMA 3} & \textbf{Claude 3} & \textbf{Gemini Pro} & \textbf{GPT-4} & \textbf{Canonical Order} & \textbf{Min-K\%} & \textbf{Token Overlap} & \textbf{WPQ} \\ \midrule
MMLU & \textcolor{teal}{\cmark} & \textcolor{teal}{\cmark} & \textcolor{teal}{\cmark} & \textcolor{teal}{\cmark} & \textcolor{teal}{\cmark} & \textcolor{purple}{\xmark} & \textcolor{purple}{\xmark} & \textcolor{purple}{\xmark} \\
AGIEval (English) & \textcolor{teal}{\cmark} & \textcolor{purple}{\xmark} & \textcolor{purple}{\xmark} & \textcolor{purple}{\xmark} & \textcolor{purple}{\xmark} & \textcolor{purple}{\xmark} & \textcolor{purple}{\xmark} & \textcolor{purple}{\xmark} \\
CommonSenseQA & \textcolor{teal}{\cmark} & \textcolor{purple}{\xmark} & \textcolor{purple}{\xmark} & \textcolor{purple}{\xmark} & \textcolor{purple}{\xmark} & \textcolor{teal}{\cmark} & \textcolor{purple}{\xmark} & \textcolor{purple}{\xmark} \\
Winogrande & \textcolor{teal}{\cmark} & \textcolor{teal}{\cmark} & \textcolor{purple}{\xmark} & \textcolor{teal}{\cmark} & \textcolor{purple}{\xmark} & \textcolor{purple}{\xmark} & \textcolor{purple}{\xmark} & \textcolor{purple}{\xmark} \\
BIG-Bench Hard & \textcolor{teal}{\cmark} & \textcolor{teal}{\cmark} & \textcolor{teal}{\cmark} & \textcolor{teal}{\cmark} & \textcolor{purple}{\xmark} & \textcolor{purple}{\xmark} & \textcolor{purple}{\xmark} & \textcolor{purple}{\xmark} \\
ARC-Challenge & \textcolor{teal}{\cmark} & \textcolor{teal}{\cmark} & \textcolor{purple}{\xmark} & \textcolor{teal}{\cmark} & \textcolor{purple}{\xmark} & \textcolor{purple}{\xmark} & \textcolor{purple}{\xmark} & \textcolor{purple}{\xmark} \\
TriviaQA-WIKI & \textcolor{teal}{\cmark} & \textcolor{purple}{\xmark} & \textcolor{purple}{\xmark} & \textcolor{purple}{\xmark} & \textcolor{purple}{\xmark} & \textcolor{purple}{\xmark} & \textcolor{purple}{\xmark} & \textcolor{purple}{\xmark} \\
SQuAD & \textcolor{teal}{\cmark} & \textcolor{purple}{\xmark} & \textcolor{purple}{\xmark} & \textcolor{purple}{\xmark} & \textcolor{purple}{\xmark} & \textcolor{purple}{\xmark} & \textcolor{purple}{\xmark} & \textcolor{purple}{\xmark} \\
QuAC & \textcolor{teal}{\cmark} & \textcolor{purple}{\xmark} & \textcolor{purple}{\xmark} & \textcolor{purple}{\xmark} & \textcolor{purple}{\xmark} & \textcolor{purple}{\xmark} & \textcolor{purple}{\xmark} & \textcolor{purple}{\xmark} \\
BoolQ & \textcolor{teal}{\cmark} & \textcolor{purple}{\xmark} & \textcolor{purple}{\xmark} & \textcolor{purple}{\xmark} & \textcolor{teal}{\cmark} & \textcolor{teal}{\cmark} & \textcolor{purple}{\xmark} & \textcolor{purple}{\xmark} \\
DROP & \textcolor{teal}{\cmark} & \textcolor{teal}{\cmark} & \textcolor{teal}{\cmark} & \textcolor{teal}{\cmark} & \textcolor{purple}{\xmark} & \textcolor{purple}{\xmark} & \textcolor{purple}{\xmark} & \textcolor{teal}{\cmark} \\
GPQA & \textcolor{teal}{\cmark} & \textcolor{teal}{\cmark} & \textcolor{purple}{\xmark} & \textcolor{teal}{\cmark} & \textcolor{purple}{\xmark} & \textcolor{purple}{\xmark} & \textcolor{purple}{\xmark} & \textcolor{purple}{\xmark} \\
HumanEval & \textcolor{teal}{\cmark} & \textcolor{teal}{\cmark} & \textcolor{teal}{\cmark} & \textcolor{teal}{\cmark} & \textcolor{purple}{\xmark} & \textcolor{purple}{\xmark} & \textcolor{purple}{\xmark} & \textcolor{teal}{\cmark} \\
GSM8K & \textcolor{teal}{\cmark} & \textcolor{teal}{\cmark} & \textcolor{teal}{\cmark} & \textcolor{teal}{\cmark} & \textcolor{purple}{\xmark} & \textcolor{purple}{\xmark} & \textcolor{purple}{\xmark} & \textcolor{teal}{\cmark} \\
MATH & \textcolor{teal}{\cmark} & \textcolor{teal}{\cmark} & \textcolor{teal}{\cmark} & \textcolor{teal}{\cmark} & \textcolor{purple}{\xmark} & \textcolor{purple}{\xmark} & \textcolor{purple}{\xmark} & \textcolor{purple}{\xmark} \\
MGSM & \textcolor{purple}{\xmark} & \textcolor{teal}{\cmark} & \textcolor{teal}{\cmark} & \textcolor{teal}{\cmark} & \textcolor{purple}{\xmark} & \textcolor{purple}{\xmark} & \textcolor{purple}{\xmark} & \textcolor{purple}{\xmark} \\
HellaSwag & \textcolor{purple}{\xmark} & \textcolor{teal}{\cmark} & \textcolor{teal}{\cmark} & \textcolor{teal}{\cmark} & \textcolor{teal}{\cmark} & \textcolor{purple}{\xmark} & \textcolor{purple}{\xmark} & \textcolor{purple}{\xmark} \\
PubMedQA & \textcolor{purple}{\xmark} & \textcolor{teal}{\cmark} & \textcolor{purple}{\xmark} & \textcolor{teal}{\cmark} & \textcolor{purple}{\xmark} & \textcolor{purple}{\xmark} & \textcolor{purple}{\xmark} & \textcolor{purple}{\xmark} \\
RACE-H & \textcolor{purple}{\xmark} & \textcolor{teal}{\cmark} & \textcolor{purple}{\xmark} & \textcolor{purple}{\xmark} & \textcolor{purple}{\xmark} & \textcolor{purple}{\xmark} & \textcolor{purple}{\xmark} & \textcolor{purple}{\xmark} \\
APPS & \textcolor{purple}{\xmark} & \textcolor{teal}{\cmark} & \textcolor{purple}{\xmark} & \textcolor{purple}{\xmark} & \textcolor{purple}{\xmark} & \textcolor{purple}{\xmark} & \textcolor{purple}{\xmark} & \textcolor{purple}{\xmark} \\
MBPP & \textcolor{purple}{\xmark} & \textcolor{teal}{\cmark} & \textcolor{purple}{\xmark} & \textcolor{purple}{\xmark} & \textcolor{purple}{\xmark} & \textcolor{purple}{\xmark} & \textcolor{purple}{\xmark} & \textcolor{purple}{\xmark} \\
Natural2Code & \textcolor{purple}{\xmark} & \textcolor{purple}{\xmark} & \textcolor{teal}{\cmark} & \textcolor{purple}{\xmark} & \textcolor{purple}{\xmark} & \textcolor{purple}{\xmark} & \textcolor{purple}{\xmark} & \textcolor{purple}{\xmark} \\
WMT23 & \textcolor{purple}{\xmark} & \textcolor{purple}{\xmark} & \textcolor{teal}{\cmark} & \textcolor{teal}{\cmark} & \textcolor{purple}{\xmark} & \textcolor{purple}{\xmark} & \textcolor{purple}{\xmark} & \textcolor{purple}{\xmark} \\
RTE & \textcolor{purple}{\xmark} & \textcolor{purple}{\xmark} & \textcolor{purple}{\xmark} & \textcolor{purple}{\xmark} & \textcolor{purple}{\xmark} & \textcolor{purple}{\xmark} & \textcolor{teal}{\cmark} & \textcolor{teal}{\cmark} \\
WNLI & \textcolor{purple}{\xmark} & \textcolor{purple}{\xmark} & \textcolor{purple}{\xmark} & \textcolor{purple}{\xmark} & \textcolor{purple}{\xmark} & \textcolor{purple}{\xmark} & \textcolor{teal}{\cmark} & \textcolor{teal}{\cmark} \\
AG News & \textcolor{purple}{\xmark} & \textcolor{purple}{\xmark} & \textcolor{purple}{\xmark} & \textcolor{purple}{\xmark} & \textcolor{purple}{\xmark} & \textcolor{purple}{\xmark} & \textcolor{teal}{\cmark} & \textcolor{teal}{\cmark} \\
MeetingBank & \textcolor{purple}{\xmark} & \textcolor{purple}{\xmark} & \textcolor{purple}{\xmark} & \textcolor{purple}{\xmark} & \textcolor{purple}{\xmark} & \textcolor{purple}{\xmark} & \textcolor{purple}{\xmark} & \textcolor{teal}{\cmark} \\
AuTexTification & \textcolor{purple}{\xmark} & \textcolor{purple}{\xmark} & \textcolor{purple}{\xmark} & \textcolor{purple}{\xmark} & \textcolor{purple}{\xmark} & \textcolor{purple}{\xmark} & \textcolor{purple}{\xmark} & \textcolor{teal}{\cmark} \\
IMDB & \textcolor{purple}{\xmark} & \textcolor{purple}{\xmark} & \textcolor{purple}{\xmark} & \textcolor{purple}{\xmark} & \textcolor{purple}{\xmark} & \textcolor{teal}{\cmark} & \textcolor{teal}{\cmark} & \textcolor{teal}{\cmark} \\
Yelp Full Reviews & \textcolor{purple}{\xmark} & \textcolor{purple}{\xmark} & \textcolor{purple}{\xmark} & \textcolor{purple}{\xmark} & \textcolor{purple}{\xmark} & \textcolor{purple}{\xmark} & \textcolor{teal}{\cmark} & \textcolor{teal}{\cmark} \\
SAMSum & \textcolor{purple}{\xmark} & \textcolor{purple}{\xmark} & \textcolor{purple}{\xmark} & \textcolor{purple}{\xmark} & \textcolor{purple}{\xmark} & \textcolor{purple}{\xmark} & \textcolor{teal}{\cmark} & \textcolor{teal}{\cmark} \\
XSum & \textcolor{purple}{\xmark} & \textcolor{purple}{\xmark} & \textcolor{purple}{\xmark} & \textcolor{purple}{\xmark} & \textcolor{purple}{\xmark} & \textcolor{purple}{\xmark} & \textcolor{teal}{\cmark} & \textcolor{teal}{\cmark} \\
OpenbookQA & \textcolor{purple}{\xmark} & \textcolor{purple}{\xmark} & \textcolor{purple}{\xmark} & \textcolor{purple}{\xmark} & \textcolor{teal}{\cmark} & \textcolor{purple}{\xmark} & \textcolor{purple}{\xmark} & \textcolor{purple}{\xmark} \\
MNLI & \textcolor{purple}{\xmark} & \textcolor{purple}{\xmark} & \textcolor{purple}{\xmark} & \textcolor{purple}{\xmark} & \textcolor{teal}{\cmark} & \textcolor{purple}{\xmark} & \textcolor{purple}{\xmark} & \textcolor{purple}{\xmark} \\
TruthfulQA & \textcolor{purple}{\xmark} & \textcolor{purple}{\xmark} & \textcolor{purple}{\xmark} & \textcolor{purple}{\xmark} & \textcolor{teal}{\cmark} & \textcolor{teal}{\cmark} & \textcolor{purple}{\xmark} & \textcolor{purple}{\xmark} \\
Natural Questions & \textcolor{purple}{\xmark} & \textcolor{purple}{\xmark} & \textcolor{purple}{\xmark} & \textcolor{purple}{\xmark} & \textcolor{teal}{\cmark} & \textcolor{purple}{\xmark} & \textcolor{purple}{\xmark} & \textcolor{purple}{\xmark} \\
PIQA & \textcolor{purple}{\xmark} & \textcolor{purple}{\xmark} & \textcolor{purple}{\xmark} & \textcolor{purple}{\xmark} & \textcolor{teal}{\cmark} & \textcolor{purple}{\xmark} & \textcolor{purple}{\xmark} & \textcolor{purple}{\xmark} \\
Books3 & \textcolor{purple}{\xmark} & \textcolor{purple}{\xmark} & \textcolor{purple}{\xmark} & \textcolor{purple}{\xmark} & \textcolor{purple}{\xmark} & \textcolor{teal}{\cmark} & \textcolor{purple}{\xmark} & \textcolor{purple}{\xmark} \\ \bottomrule
\end{tabular}%
}
\caption{The dataset mismatch between the challenging benchmarks frequently evaluated by SOTA LLMs and the ones used for validation of contamination detection methods.}
\label{tab:dataset-mismatch}
\end{table*}

\end{document}